\documentclass{article} 
\usepackage{preprint,times}

\usepackage{amsmath,amsfonts,bm}

\def\eqref#1{equation~\ref{#1}}

\def\1{\bm{1}}

\DeclareMathAlphabet{\mathsfit}{\encodingdefault}{\sfdefault}{m}{sl}
\SetMathAlphabet{\mathsfit}{bold}{\encodingdefault}{\sfdefault}{bx}{n}

\usepackage{latexsym}
\usepackage[T1]{fontenc}
\usepackage[utf8]{inputenc}
\usepackage{microtype}
\usepackage{inconsolata}
\usepackage{graphicx}
\usepackage{booktabs}
\usepackage{multirow}
\usepackage{array}
\usepackage{amsmath}
\usepackage{amssymb}
\usepackage{xcolor}
\usepackage{xspace}
\usepackage{tikz}
\usetikzlibrary{arrows.meta, positioning, shapes.geometric, decorations.pathmorphing, shapes.symbols}

\usepackage{hyperref}
\usepackage{url}
\usepackage{xurl}
\hypersetup{hidelinks,
  pdftitle={KLineage: Recovering the Missing When of Kernel Optimization by Deoptimizing Experts},
  pdfauthor={Shuoming Zhang, Qiuchu Yu, Ruiyuan Xu, Chenjing Zhang, Junjie Peng, Zhicheng Xie, Yangyu Zhang, Xiyu Shi, Ying Liu, Guangli Li, Xiaobing Feng, Huimin Cui, Xingjun Zhang, Jiacheng Zhao}}

\newcommand{\sys}{\textsc{KLineage}\xspace}

\title{\sys: Recovering the Missing \emph{When} of \\ Kernel Optimization by Deoptimizing Experts}

\author{%
\begin{minipage}[t]{\dimexpr\textwidth-2\tabcolsep\relax}
\centering
Shuoming Zhang\textsuperscript{1,2,*} \quad Qiuchu Yu\textsuperscript{1,2,*} \quad Ruiyuan Xu\textsuperscript{1,2} \quad Chenjing Zhang\textsuperscript{3} \\
Junjie Peng\textsuperscript{3} \quad Zhicheng Xie\textsuperscript{3} \quad Yangyu Zhang\textsuperscript{1,2} \quad Xiyu Shi\textsuperscript{1} \\
Ying Liu\textsuperscript{1,2} \quad Guangli Li\textsuperscript{1,2} \quad Xiaobing Feng\textsuperscript{1,2} \quad Huimin Cui\textsuperscript{1,2} \\
Xingjun Zhang\textsuperscript{3} \quad Jiacheng Zhao\textsuperscript{1,2,\dag} \\[0.5em]
{\normalfont\small
\textsuperscript{1}State Key Laboratory of Processors, Institute of Computing Technology\\
Chinese Academy of Sciences, Beijing, China\\
\textsuperscript{2}University of Chinese Academy of Sciences, Beijing, China\\
\textsuperscript{3}Xi'an Jiaotong University, Xi'an, China}
\end{minipage}%
}

\begin{document}
\raggedbottom

\maketitle
\lhead{}
\renewcommand{\headrulewidth}{0pt}
\begingroup
\renewcommand{\thefootnote}{}
\footnotetext{Preprint.}
\renewcommand{\thefootnote}{*}
\footnotetext{Equal contribution.}
\renewcommand{\thefootnote}{\dag}
\footnotetext{Corresponding to: \texttt{zhaojiacheng@ict.ac.cn}.}
\endgroup

\begin{abstract}

LLM-based agents are increasingly used to generate GPU kernels, but they often struggle to determine when an optimization is sound because its required code state and dependencies are implicit in expert implementations. We introduce \sys, which learns this missing ``when'' knowledge from expert kernels: instead of relying on forward rollouts, \sys walks expert implementations backward through validation-gated simplifications and reverses each accepted step into a reusable optimization skill. Each skill records not only the optimization intent, but also when to apply the optimization technique, including where it applies in code, what conditions made it valid, what effect it has, and what failures its assumptions avoid. A downstream LLM materializes these skills on new code surfaces under the same compile/correctness/profile gate. This guidance on when to apply each optimization can help downstream models to generate higher-performance kernels.    
On five expert workloads across two NVIDIA architectures, these lineage-derived skills serve as an effective optimization curriculum, exceeding recent memory-based LLM-kernel baselines in both final kernel quality and optimization efficiency under the same fixed budget. 
We also demonstrate that the \sys framework extends beyond NVIDIA GPUs to Ascend NPUs. Our code is publicly available at {\def\UrlBreaks{\do\/\do\-}\url{https://github.com/ict-agent/klineage}}.

\end{abstract}

\section{Introduction}
\label{sec:intro}
LLM-based agents are becoming a practical interface for GPU-kernel generation: they can draft an implementation, run it, inspect compiler or execution feedback, and iterate. Benchmarks such as KernelBench~\citep{kernelbench} and TritonBench~\citep{tritonbench} have made this setting measurable, while recent agents improve kernels through feedback, search, post-training, or skill memory~\citep{gpu_kernel_scientist,geak,sakana_cuda_engineer,kernel_smith,kernel_skill,adaexplore}. The setting is also increasingly multi-surface: agents may write CUDA, Triton, CuTe/CUTLASS-style templates, TileLang programs, or a restricted kernel DSL, each exposing different scheduling knobs and failure modes~\citep{tritonbench,geak,cutegen,tritonforge,hari2026improving,chu2025gpu,qu2026two}. However, prior studies document a gap between attempting an optimization and applying it effectively. KernelBench reports that providing optimization examples can encourage more aggressive transformations while increasing execution failures~\citep{kernelbench}. 

The problem is often not that the model lacks the names of useful optimizations. It can suggest tiling, shared-memory staging, vectorized loads, software pipelining, or hardware intrinsics. What is missing is the condition under which each optimization is valid and useful. Vectorized loads require alignment and valid tail handling; pipelining assumes staged data movement; matrix intrinsics require compatible tile shapes, layouts, and hardware support. A forward-search agent may therefore apply a familiar optimization to the wrong state, producing a compile error, a correctness failure, or a slower kernel. Pretraining captures much of the \emph{what} of optimization, but not reliably the \emph{when}.

We treat this missing ``when'' knowledge as the object to be learned. The memory item we need is not a free-form hint such as ``use vectorized loads'', but a skill in a narrow sense: a verified, code-anchored specification of an optimization intent, including where the intent acts on a code or IR surface, what carrier can guide implementation, the context in which it has been observed to work, the effect it tends to have, and the risks when its assumptions are violated. Existing kernel-agent memories expose parts of this signal through failure rules, profile-state knowledge bases, or expert-skill snippets~\citep{adaexplore,ksearch,kernelblaster,kernel_skill}, while post-training approaches may absorb it into model weights~\citep{kernel_smith,kevin_rl_kernel,cuda_l1}. 
These memories are useful, but when they are built from forward rollouts, their coverage is limited by the trajectories the agent was able to discover. The optimization experience they accumulate depends on what the backbone model can propose and successfully implement. When high performance requires unfamiliar combinations of hardware features and optimization steps, forward rollouts can remain concentrated on familiar implementation patterns, leaving the relevant code states and dependencies unexplored within a fixed budget. Expert kernels supply successful implementations directly, giving \sys a concrete basis for recovering optimization paths and applicability conditions that the agent may fail to discover on its own. Our goal is to externalize the missing ``when'' knowledge from expert artifacts instead.

We introduce \sys, a framework for inducing reusable skills from expert GPU-kernel lineages. The key observation is that expert kernels already contain the missing conditional knowledge, but only implicitly. 
Expert kernels are developed through a sequence of interdependent optimization decisions involving tiling, vectorization, parameter selection, and hardware features. However, the final code typically leaves the intermediate code states, optimization order, and conditions that make each step valid implicit.\sys recovers these decisions by walking an expert kernel backward through named, validation-gated deoptimization steps. Each accepted simplification is stored as its inverse forward transition, yielding a curriculum from a naive implementation toward the expert kernel; the forward transitions observed across experts become evidence for reusable skills. To our knowledge, \sys is the first LLM-kernel memory constructed by deoptimizing expert implementations rather than accumulating forward-rollout states.


At reuse time, \sys retrieves skills by scope. 
Given a new target, it retrieves skills along the target case, surface language, platform, and already-applied optimizations. The target state need not be naive: \sys can start from any current kernel attempt and use the retrieved lineage to complete missing intents or repair mis-materialized ones. It then lets a downstream LLM materialize selected skills on the target code surface using their anchors and carriers.
This keeps the LLM in the role where it is useful---instantiating a validated code-anchored intent and repairing local mismatches under feedback---while the memory constrains it with validated preconditions and observed risks.

We evaluate \sys{} on two NVIDIA platforms, an H100 and an RTX PRO 6000, and on an Ascend 910B, inducing one library on each from five expert kernels. On those kernels \sys{} is correct on all ten workload--platform pairs at $1.12\times$ over the vendor references, where AdaExplore~\citep{adaexplore} and AccelOpt~\citep{accelopt} reach $0.25\times$ and $0.54\times$ and return nothing correct on four and two of them. An agent handed the same expert kernels as source recovers individual optimizations but not the combinations they belong to, and its kernels run up to $105\times$ slower. On five new operators per platform that contribute no skills, \sys{} reaches $1.73\times$ on H100 and $1.41\times$ on Ascend over that same agent; on the $100$ Level~2 problems of KernelBench~\citep{kernelbench}, ported to AscendC as the Fusion category of MultiKernelBench~\citep{multikernelbench}, mean speedup rises from $1.05\times$ to $1.32\times$ against \texttt{torch.compile} on NVIDIA and from $0.63\times$ to $2.09\times$ against eager PyTorch on Ascend. Conditioned skills therefore make expert optimization knowledge reusable well beyond the kernels it was extracted from.

\paragraph{Contributions.}
This paper introduces \sys{} and makes three technical contributions:
\begin{itemize}
    \item \textbf{Conditional skills that encode when to optimize.} We make the ``when'' of optimization explicit by pairing each optimization intent with the conditions under which it is valid and useful. Each skill includes code anchors and implementation carriers to guide materialization, together with effects, risks, verification evidence, and scope over workloads, languages, platforms, and prior optimization states.

    \item \textbf{Guided deoptimization for constructing conditional skills.} We walk expert kernels backward through validation-gated simplifications and reconstruct accepted steps as validated forward transitions, forming an executable curriculum from naive to expert. We then lift these transitions into conditional skills, using the intermediate code states, optimization dependencies, and validation outcomes as evidence for their applicability conditions.

    \item \textbf{Using conditional skills to guide when to optimize.} We retrieve conditional skills according to the target kernel's current state and context, using their applicability conditions to guide which optimizations to apply at each stage. Only skills admitted through held-out roundtrip validation are reused; a downstream LLM materializes the selected skills into optimized kernels under a compile/correctness/profile gate.
\end{itemize}

\section{Related Work}
\label{sec:related}

\begin{figure*}[t]
\centering
\includegraphics[width=\textwidth]{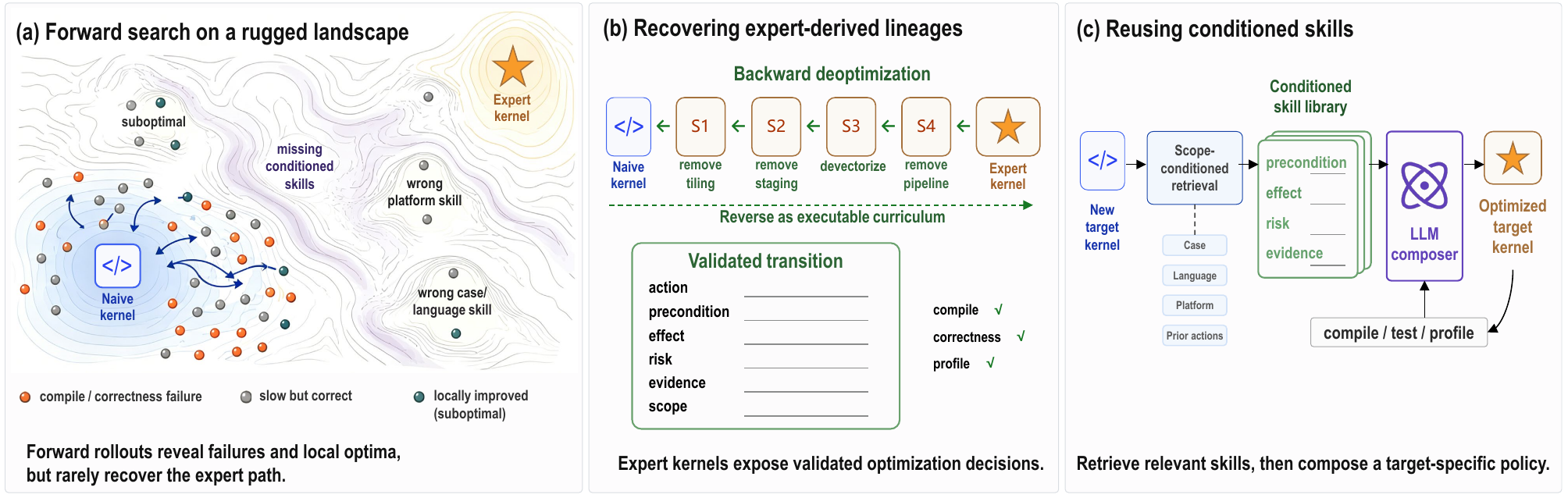}
\caption{\small
\textbf{Motivation of \sys.}
\textbf{(a)} Forward search often knows which optimizations to try but not when their preconditions hold.
\textbf{(b)} \sys walks expert kernels backward to recover validated forward transitions from simpler states toward expert states.
\textbf{(c)} The recovered skills are reused only as code-anchored candidates, with compile/test/profile gates deciding admission on new targets.
}
\label{fig:lineage-teaser}
\end{figure*}

\paragraph{LLM agents for GPU kernel optimization.}
Recent benchmarks such as KernelBench~\cite{kernelbench}, TritonBench~\cite{tritonbench}, and production-oriented FastKernels~\citep{fastkernels}, together with kernel substrates such as ThunderKittens~\cite{thunderkittens}, have made GPU-kernel generation a common testbed for LLM agents. Recent systems improve kernels through compiler/profile feedback, multi-agent refinement, DSL or minimal-program interfaces, static-analysis guidance, post-training, and search over CUDA, Triton, CuTe/CUTLASS-style, and related surfaces~\citep{gpu_kernel_scientist,geak,sakana_cuda_engineer,qimeng_kernel,astra,cutegen,tritonforge,hari2026improving,chu2025gpu,qu2026two,lou2024automatic,keet,kevin_rl_kernel,cuda_l1,autotriton}. KernelFoundry~\citep{kernelfoundry}, for example, uses quality-diversity search, prompt evolution, and template tuning to explore kernels under hardware profiling feedback, while AKO4ALL~\citep{ako4all} drives an agent through a fixed optimization protocol and Atrex Kernel Agent~\citep{atrex} works a profile-driven loop over operators sampled from production inference traces. These systems establish realistic compile/profile loops, but they largely acquire optimization knowledge by moving forward from an initial program.

\paragraph{Reusable experience and memory.}
Closest to our work are kernel agents that reuse optimization experience: AdaExplore~\cite{adaexplore} converts failures into textual rules; AccelOpt~\citep{accelopt} summarizes a run's evolving state; K-Search~\citep{ksearch} co-evolves kernels with an intrinsic world model; KernelSkill~\cite{kernel_skill} organizes expert-written skills; and KernelBlaster~\citep{kernelblaster} stores forward-rollout experience in a profile-centric memory. CuBridge~\citep{cubridge} is also related in using LLMs to understand and reconstruct expert attention kernels. \sys differs in how the memory is constructed and represented: to our knowledge, it is the first LLM-kernel memory construction method that automatically derives reusable lineages by deoptimizing expert implementations, and each skill records code/IR anchors, carriers, preconditions, effects, risks, evidence, and scope rather than only a forward trace, textual rule, or pre-existing snippet.

\section{Method}
\label{sec:method}

\sys constructs reusable, code-anchored optimization memory from expert evidence. Guided deoptimization walks an expert kernel backward but stores accepted steps as inverse forward transitions (\autoref{sec:method-deopt}); a lift turns recurring transitions into candidate skill hypotheses (\autoref{sec:method-extract}); held-out roundtrip materialization decides which hypotheses become reusable skills (\autoref{sec:method-use}). 

\subsection{Motivation}
\label{sec:method-formulation}

\begin{figure*}[t]
\centering
\includegraphics[width=\textwidth]{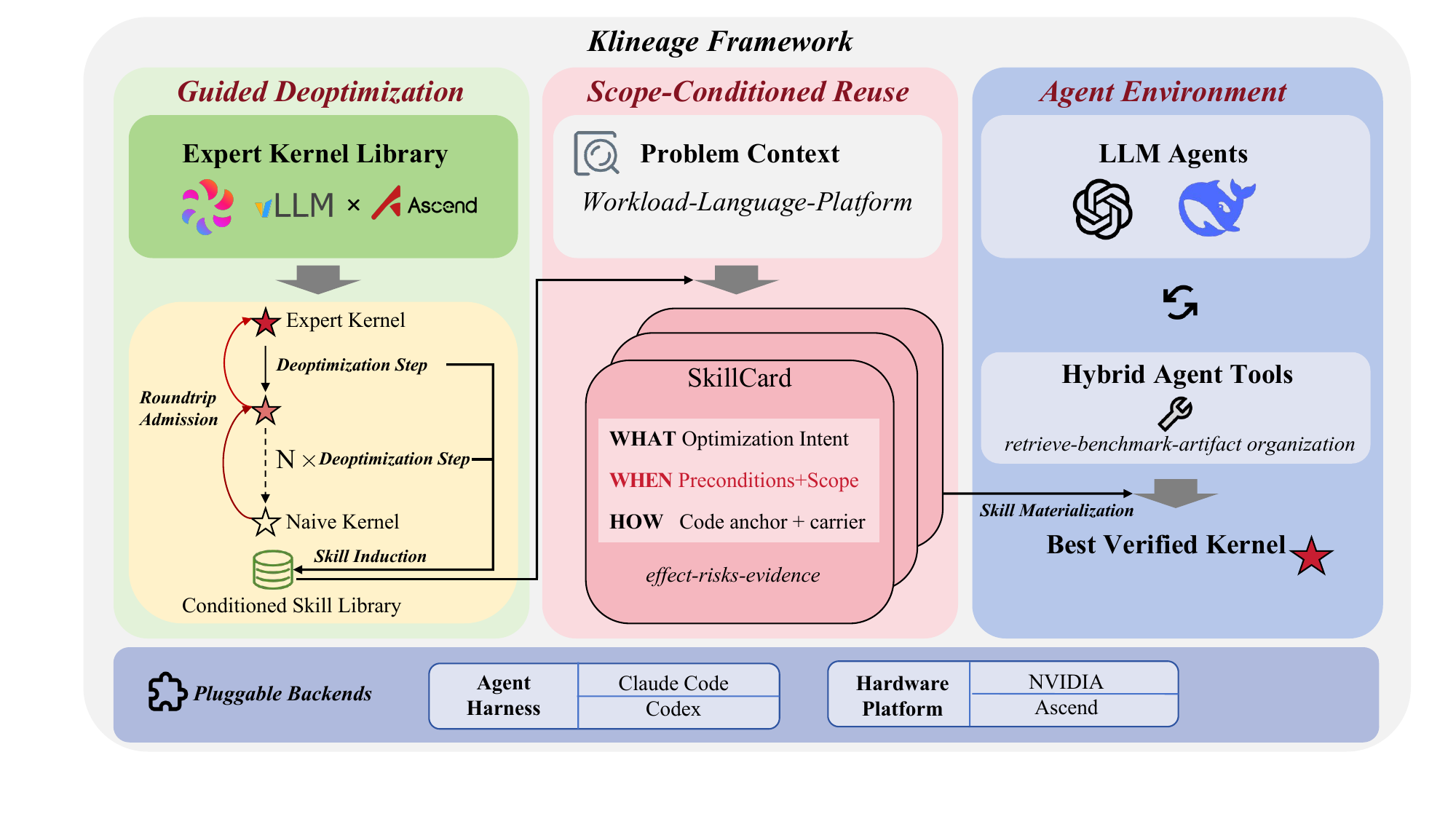}
\caption{\small Overview of \sys{}.}
\label{fig:method-overview}
\end{figure*}

As illustrated in \autoref{fig:lineage-teaser}, expert kernels entangle many compatible optimization decisions, and walking them backward under validation exposes each decision as an explicit forward transition. This direction is more natural in practice because high-quality expert kernels in existing repositories are often not paired with corresponding naive implementations. We make this construction precise below.

\sys represents this evidence at three levels as shown in ~\autoref{fig:lineage-teaser}(b). A \emph{backward deoptimization step} $b_i : K_i \rightarrow K_{i-1}$ removes or weakens one optimization and records the edit, locus, and validation evidence for the simplified predecessor. \sys stores the inverse $\tau_i = b_i^{-1}:K_{i-1}\rightarrow K_i$ as a \emph{forward concrete transition}: a validated before/after edit from a simpler state toward the expert state. A \emph{lineage} is the resulting forward sequence
\[
L = (K_0, \tau_1, K_1, \ldots, \tau_n, K_n = K^*),
\]
an executable curriculum from naive to expert, not a claim about the expert's historical writing path.

Concrete transitions are still tied to one source kernel. \sys lifts them into an \emph{optimization intent}: the abstract pass family being expressed, such as staging reused tiles, vectorizing aligned contiguous loads, or exposing tunable parameters. A \emph{candidate skill hypothesis} $\hat{\sigma}$ makes the intent actionable by adding a code/IR anchor, a carrier, conditions, and evidence:
\[
\begin{aligned}
\hat{\sigma} = \big(&\iota,\, \mathrm{anchor},\, \mathrm{carrier},\, \mathrm{pre},\\
                   &\mathrm{effect},\, \mathrm{evidence},\, \mathrm{risk},\,
                    \mathrm{scope},\, \mathrm{ver}\big),
\end{aligned}
\]
where $\iota$ is the intent, $\mathrm{anchor}$ identifies the code or IR surface on which it acts, and $\mathrm{carrier}$ is an actionable but not necessarily executable representation such as a diff sketch, pseudocode, annotated kernel snippet, schedule fragment, or natural-language code annotation. The other fields record preconditions, expected effects, evidence, risks, scope over case/language/platform/prior-action context, and a verification log $\mathrm{ver}$ (\autoref{sec:method-extract}). A hypothesis becomes an admitted \emph{skill} only after at least one trial succeeds.
At reuse time, an agent \emph{materializes} the skill on a target code surface; the candidate is accepted only if it passes the compile/correctness/profile gate.

\subsection{Guided Deoptimization}
\label{sec:method-deopt}

Starting from $K_n = K^*$, \sys repeatedly proposes a simplification (e.g., remove software pipelining, devectorize, scalarize an intrinsic), patches the kernel with an LLM rewriter, and accepts the patch only if it passes validation. The inverse of an accepted simplification becomes a forward transition in the lineage; rejected simplifications become precondition-violation evidence for later $\mathrm{risk}$ fields.

\paragraph{Operationalizing $\tau_i = b_i^{-1}$.}
The forward transition $\tau_i$ is not a textual reversal of the diff that produced $b_i$. The LLM rewriter re-derives a forward edit on $K_{i-1}$ functionally equivalent to $K_i$, given the action category and locus from the backward step. The candidate forward edit is then run through the compile/correctness/profile gate; only edits whose output kernel matches $K_i$'s validation status are stored as $\tau_i$. This makes $\tau_i$ an executable forward primitive grounded in $K_{i-1}$'s state, not a syntactic inversion that may not even compile.

During induction we group accepted backward steps by action category to anchor the lift, over an open vocabulary rather than a closed set of actions. Multiple intents may be lifted from one category, the same intent may admit different carriers in different languages, and any partial order over actions is treated as a soft prior: if a simplification violates the order but still passes validation, its inverse transition is kept.

\subsection{From Transitions to Skills}
\label{sec:method-extract}

The library is induced from forward concrete transitions. Transitions sharing an action category and similar locus / effect signatures are aggregated, and an LLM receives their forward diffs, loci, profile deltas, and local code context, then lifts them into an intent $\iota$, an anchor, a carrier, and a portable $(\mathrm{pre}, \mathrm{effect}, \mathrm{risk})$ tuple. The intent describes the manipulated code construct and exploited platform property; the anchor and carrier keep it actionable.

\paragraph{Intent, carrier, and materialization.}
The lift separates the general idea from its realization. The intent names the optimization, the carrier gives it a code-bearing form, and materialization instantiates the admitted skill on a target language and code surface. One intent can therefore have different realizations across cases, languages, and platforms.

\paragraph{Roundtrip admission.}
An induced $\hat{\sigma}$ starts as a hypothesis. Admission requires at least one held-out roundtrip in which an agent, given only $\hat{\sigma}$, materializes it on a fresh state and reproduces its expected effect; success is recorded in $\mathrm{ver}$. Only admitted skills $\sigma \in \mathcal{S}$ are reused.

\subsection{Scope-Conditioned Reuse}
\label{sec:method-use}

\sys supplies a verified, scope-conditioned candidate pool rather than a new search policy. Given a target $T$ and a root submission, the downstream agent selects admitted skills, materializes them on the target code surface, and submits candidates under the same fixed-budget compile/correctness/profile loop used by all baselines. In the online prompt, each retrieved SkillCard is serialized as the fields in $\hat{\sigma}$---intent, anchor, carrier, precondition, effect, evidence, risk, scope, and verification log---with the carrier copied verbatim as the code-bearing hint to instantiate, not as an executable tool. The root submission may be a compiler baseline, a partially optimized agent attempt, or a previous best kernel.
To bound the candidate pool, \sys retrieves with a multi-faceted retriever: given $T = (c_T, l_T, p_T, d_T)$, it returns the union of per-dimension top-$k$ skills under similarities over $(\mathcal{C}, \mathcal{L}, \mathcal{P}, \mathcal{D})$, using each skill's verified rather than declared scope.

\subsection{\sys Framework}

Building on these ideas, we implement \sys as an agent framework for GPU kernel programming. As shown in \autoref{fig:method-overview}, \sys induces conditioned skills from expert-written kernels such as FlashInfer~\cite{ye2025flashinfer} through guided deoptimization. Each skill records not only an optimization, but also the conditions under which it applies, such as data layout, hardware features, memory placement, and pipeline schedule. Given a new workload with the specified language and target platform, \sys guides LLM agents to retrieve and materialize skills whose conditions match the current context. We also provide hybrid agent tools for LLM Agents in the environment, including shared interfaces for skill retrieval, kernel artifact organization, and benchmarking, together with platform-specific profiling tools such as \texttt{ncu} and \texttt{msprof}. \sys{} also provides a unified abstraction for pluggable integration of different backends and agent harnesses. This design makes agent trajectories and generated artifacts traceable while leaving agents free to profile kernels, diagnose bottlenecks, and explore platform-specific optimizations.

\section{Evaluation}
\label{sec:experiments}

We structure our evaluation around four research questions:

\noindent\textbf{RQ1:} How do kernels generated by \sys{} compare with
those produced by existing memory-based agents and with expert
implementations?

\noindent\textbf{RQ2:} Does a \sys improve kernel generation over providing expert source code directly?

\noindent\textbf{RQ3:} How well does \sys{} generalize to non-NVIDIA accelerators?

\noindent\textbf{RQ4:} How effectively can the induced skills be transfered to other operators and broader benchmarks?

\subsection{Setup}
\label{sec:exp-setup}

\paragraph{Platforms and languages.}
We evaluate on NVIDIA H100 PCIe (SM90), NVIDIA RTX PRO 6000 Blackwell (SM120), and Ascend 910B. On Nvidia, we use PyTorch 2.11.0+cu130, FlashInfer 0.6.11, TileLang 0.1.8, and FLA 0.5.0. FLA replaces FlashInfer as the GDN reference on SM120. On Ascend, we use CANN 9.1.0 and Torch-NPU 2.10.0. We build a separate skill library for each platform and record the verified architecture in each skill's scope. The target languages are CUDA C on NVIDIA and AscendC on Ascend.

\paragraph{Expert kernels and benchmarks.}
We select five expert kernels per platform, covering GEMM, Conv2d, FMHA, GDN, and Top-K. These operators are common in production, and their expert implementations contain a range of optimization techniques. Appendix~\ref{app:workload-detail} lists the workload shapes, data types, and reference implementations. The transfer targets and benchmark problems do not contribute to skill induction. We evaluate the transferability of the induced skills using 100 KernelBench Level~2 problems on NVIDIA GPUs and 100 problems from MultiKernelBench's Fusion category on Ascend NPUs.

\paragraph{Agents and budgets.}
We use Claude Code with Claude Opus 4.6 as the harness and LLM backbone  in \sys for offline skill induction and online kernel generation on both Nvidia and Ascend platforms. The online generation budget for these experiments is \$10 per workload; induction costs are reported separately. For operator transfer and benchmark evaluation on both platforms, we use Codex with DeepSeek-V4.1-Flash. Each transfer target has a two-hour wall-clock budget. Each problem in KernelBench or MultiKernelBench has a 30-minute wall-clock budget.
All reported costs are calculated using the corresponding model
providers' official API pricing.

\subsection{RQ1: Kernel Generation under induced Skill Library}
\label{sec:exp-induction}

We induce one library for each NVIDIA platform by the procedure of \autoref{sec:method-deopt} and \autoref{sec:method-extract}: walk each expert kernel backward under the checks, keep the reverse of every accepted step, and admit a lifted skill only once an agent handed that skill alone reproduces its effect on the deoptimized kernel.

As shown in \autoref{tab:roundtrip}, this procedure yields 52 admitted SkillCards on H100 and 36 on RTX PRO 6000 from the five expert kernels per platform listed in \autoref{tab:workload-detail}. The roundtrip admission rate is lower on H100: $72.2\%$ ($52/72$), compared with $80.0\%$ ($36/45$) on RTX PRO 6000. This difference is consistent with the more complex optimizations in the H100 experts, which use features such as distributed shared memory and warp-group tensor-core (WGMMA) instructions. Each SkillCard records the code and preconditions for applying an optimization, such as the layout requirements for WGMMA on H100. Induction is a one-time offline cost of \$131 on H100 and \$92 on RTX PRO 6000.

\begin{table}[h]
  \centering
  \small
  \setlength{\tabcolsep}{8pt}
  \begin{tabular}{@{}lccc@{}}
    \toprule
    \textbf{Platform} & \textbf{Skills in library} & \textbf{Roundtrip Admission} & \textbf{Induction cost (USD)} \\
    \midrule
    H100 (SM90)          & 52 & 52/72 & 131 \\
    RTX PRO 6000 (SM120) & 36 & 36/45 & 92 \\
    \bottomrule
  \end{tabular}
  \caption{The number of induced skills, roundtrip admission rate and induction cost of \sys{} on 2 Nvidia platforms across 5 expert kernels.}
  \label{tab:roundtrip}
\end{table}

\paragraph{What published memory-based agents reach on the same five kernels.}
Two systems already build a reusable memory for kernel optimization: AdaExplore~\citep{adaexplore}, which turns failures into rules and searches with MCTS, and AccelOpt~\citep{accelopt}, which keeps an LLM-summarized memory over the kernels it has generated. On the same five expert workloads, they reach $0.25\times$ and $0.54\times$ against the vendor reference and return no correct kernel at all on four and two of the ten workload--platform pairs; \sys{} reaches $1.12\times$ and is correct on all ten (\autoref{fig:cuda-bars}, absolute latencies in \autoref{tab:appendix-cuda-latencies}). Neither baseline converts more budget into a structurally different kernel, so we settle the comparison here and read the rest of the evaluation as \sys{} against the expert code it was induced from.

In \autoref{fig:cuda-bars}, a red $\times$ marks a workload with no correct kernel within the budget, and $\dagger$ marks the post-hoc adapter audit of AdaExplore's SM120 Conv2d kernel after excluding fallback wrappers.

\begin{figure*}[t]
\centering
\begin{tikzpicture}[x=1.05cm,y=1.13cm]
  \node[anchor=south,font=\scriptsize] at (6.0,2.35) {%
    \tikz{\draw[fill=blue!65,draw=none] (0,0) rectangle (0.30,0.075);}~\sys{}\quad
    \tikz{\draw[fill=orange!75,draw=none] (0,0) rectangle (0.30,0.075);}~AdaExplore\quad
    \tikz{\draw[fill=gray!65,draw=none] (0,0) rectangle (0.30,0.075);}~AccelOpt};
\begin{scope}
  \draw[->] (-0.1,0) -- (5.75,0);
  \draw[->] (-0.1,0) -- (-0.1,1.95);
  \draw[densely dashed] (-0.1,1) -- (5.6,1);
  \draw (-0.18,0) -- (-0.02,0);
  \node[left,inner sep=1.5pt,font=\tiny] at (-0.18,0) {0};
  \draw (-0.18,0.5) -- (-0.02,0.5);
  \node[left,inner sep=1.5pt,font=\tiny] at (-0.18,0.5) {.5};
  \draw (-0.18,1) -- (-0.02,1);
  \node[left,inner sep=1.5pt,font=\tiny] at (-0.18,1) {1};
  \draw (-0.18,1.5) -- (-0.02,1.5);
  \node[left,inner sep=1.5pt,font=\tiny] at (-0.18,1.5) {1.5};
  \node[rotate=90,anchor=south,font=\tiny] at (-0.62,0.95) {Speedup};
  \node[anchor=south,font=\scriptsize] at (2.73,1.98) {SM90 / H100};
  \draw[fill=blue!65,draw=none] (0.0,0) rectangle ++(0.26,0.99);
  \draw[fill=orange!75,draw=none] (0.3,0) rectangle ++(0.26,0.97);
  \draw[fill=gray!65,draw=none] (0.6,0) rectangle ++(0.26,0.97);
  \node[anchor=south,inner sep=0.8pt,fill=white,font=\tiny] at (0.13,1.02) {$0.99{\times}$};
  \node[anchor=north,font=\tiny] at (0.43,-0.04) {GEMM};
  \draw[fill=blue!65,draw=none] (1.15,0) rectangle ++(0.26,1.73);
  \draw[fill=orange!75,draw=none] (1.45,0) rectangle ++(0.26,0.4);
  \draw[fill=gray!65,draw=none] (1.75,0) rectangle ++(0.26,0.41);
  \node[anchor=south,inner sep=0.8pt,fill=white,font=\tiny] at (1.2799999999999998,1.76) {$1.73{\times}$};
  \node[anchor=north,font=\tiny] at (1.5799999999999998,-0.04) {Conv};
  \draw[fill=blue!65,draw=none] (2.3,0) rectangle ++(0.26,0.94);
  \node[inner sep=0pt,font=\bfseries\tiny,red!75!black] at (2.7299999999999995,0.13) {$\times$};
  \draw[fill=gray!65,draw=none] (2.9,0) rectangle ++(0.26,0.64);
  \node[anchor=south,inner sep=0.8pt,fill=white,font=\tiny] at (2.4299999999999997,0.97) {$0.94{\times}$};
  \node[anchor=north,font=\tiny] at (2.73,-0.04) {FMHA};
  \draw[fill=blue!65,draw=none] (3.4499999999999997,0) rectangle ++(0.26,1.15);
  \draw[fill=orange!75,draw=none] (3.7499999999999996,0) rectangle ++(0.26,0.02);
  \node[inner sep=0pt,font=\bfseries\tiny,red!75!black] at (4.18,0.13) {$\times$};
  \node[anchor=south,inner sep=0.8pt,fill=white,font=\tiny] at (3.5799999999999996,1.18) {$1.15{\times}$};
  \node[anchor=north,font=\tiny] at (3.88,-0.04) {GDN};
  \draw[fill=blue!65,draw=none] (4.6,0) rectangle ++(0.26,1.12);
  \draw[fill=orange!75,draw=none] (4.8999999999999995,0) rectangle ++(0.26,0.12);
  \draw[fill=gray!65,draw=none] (5.199999999999999,0) rectangle ++(0.26,0.13);
  \node[anchor=south,inner sep=0.8pt,fill=white,font=\tiny] at (4.7299999999999995,1.1500000000000001) {$1.12{\times}$};
  \node[anchor=north,font=\tiny] at (5.029999999999999,-0.04) {Top-K};
\end{scope}
\begin{scope}[xshift=6.55cm]
\begin{scope}
  \draw[->] (-0.1,0) -- (5.75,0);
  \draw[->] (-0.1,0) -- (-0.1,1.95);
  \draw[densely dashed] (-0.1,1) -- (5.6,1);
  \draw (-0.18,0) -- (-0.02,0);
  \draw (-0.18,0.5) -- (-0.02,0.5);
  \draw (-0.18,1) -- (-0.02,1);
  \draw (-0.18,1.5) -- (-0.02,1.5);
  \node[anchor=south,font=\scriptsize] at (2.73,1.98) {SM120 / RTX PRO 6000};
  \draw[fill=blue!65,draw=none] (0.0,0) rectangle ++(0.26,1.0);
  \draw[fill=orange!75,draw=none] (0.3,0) rectangle ++(0.26,0.95);
  \draw[fill=gray!65,draw=none] (0.6,0) rectangle ++(0.26,0.94);
  \node[anchor=south,inner sep=0.8pt,fill=white,font=\tiny] at (0.13,1.03) {$1.00{\times}$};
  \node[anchor=north,font=\tiny] at (0.43,-0.04) {GEMM};
  \draw[fill=blue!65,draw=none] (1.15,0) rectangle ++(0.26,1.08);
  \draw[fill=orange!75,draw=none] (1.45,0) rectangle ++(0.26,0.33);
  \draw[fill=gray!65,draw=none] (1.75,0) rectangle ++(0.26,0.31);
  \node[anchor=south,inner sep=0.8pt,fill=white,font=\tiny] at (1.2799999999999998,1.11) {$1.08{\times}$};
  \node[anchor=south,inner sep=0.8pt,fill=white,font=\tiny] at (1.5799999999999998,0.36) {$^\dagger$};
  \node[anchor=north,font=\tiny] at (1.5799999999999998,-0.04) {Conv};
  \draw[fill=blue!65,draw=none] (2.3,0) rectangle ++(0.26,0.99);
  \node[inner sep=0pt,font=\bfseries\tiny,red!75!black] at (2.7299999999999995,0.13) {$\times$};
  \draw[fill=gray!65,draw=none] (2.9,0) rectangle ++(0.26,0.88);
  \node[anchor=south,inner sep=0.8pt,fill=white,font=\tiny] at (2.4299999999999997,1.02) {$0.99{\times}$};
  \node[anchor=north,font=\tiny] at (2.73,-0.04) {FMHA};
  \draw[fill=blue!65,draw=none] (3.4499999999999997,0) rectangle ++(0.26,1.08);
  \node[inner sep=0pt,font=\bfseries\tiny,red!75!black] at (3.8799999999999994,0.13) {$\times$};
  \node[inner sep=0pt,font=\bfseries\tiny,red!75!black] at (4.18,0.13) {$\times$};
  \node[anchor=south,inner sep=0.8pt,fill=white,font=\tiny] at (3.5799999999999996,1.11) {$1.08{\times}$};
  \node[anchor=north,font=\tiny] at (3.88,-0.04) {GDN};
  \draw[fill=blue!65,draw=none] (4.6,0) rectangle ++(0.26,1.29);
  \node[inner sep=0pt,font=\bfseries\tiny,red!75!black] at (5.029999999999999,0.13) {$\times$};
  \draw[fill=gray!65,draw=none] (5.199999999999999,0) rectangle ++(0.26,0.88);
  \node[anchor=south,inner sep=0.8pt,fill=white,font=\tiny] at (4.7299999999999995,1.32) {$1.29{\times}$};
  \node[anchor=north,font=\tiny] at (5.029999999999999,-0.04) {Top-K};
\end{scope}
\end{scope}
\end{tikzpicture}
\caption{Generated CUDA-kernel speedup over the platform-specific vendor reference. }
\label{fig:cuda-bars}
\end{figure*}

\subsection{RQ2: Expert Source versus Conditioned Skills}
\label{sec:exp-source-ablation}

We replace the conditioned skill library with expert kernel sources, keeping the rest of the setup fixed and allowing iterative refinement in both settings.

Generation from expert source is slower on all ten workload--platform pairs: $5.83$--$14.48\times$ on Conv2d and $37.43$--$105.18\times$ on GDN. The gap is smaller on GEMM and Top-K, at $1.12$--$1.48\times$. Per-workload latencies are reported in \autoref{tab:expert-source-ablation} (Appendix~\ref{app:tables}).

Even after repeated refinement, agents given expert source code often converge to simpler implementations that omit complex combinations of optimizations. On Conv2d, the agent reaches the appropriate tensor-core intrinsic but fails to combine it with the required shared-memory staging, swizzling, and pipelining. On SM90 GDN, it recovers the local \texttt{cumsum} transpose but omits the subsequent WGMMA, pipelining, and TMA optimizations. In both cases, the agent recovers an individual optimization while missing the intermediate code states needed to compose it with others. These results show that conditioned skills record these prerequisites and guide the agent in building the intermediate states needed for more complex optimizations. Roundtrip admission validates individual skills so that the LLM can tackle long-horizon optimization through small, verified steps, progressively implementing expert strategies while exploring alternative implementations.

\subsection{RQ3: Generalization beyond NVIDIA GPUs}
\label{sec:exp-ascend}

On accelerators with less publicly available kernel code, coding agents may have less prior knowledge of platform-specific optimizations. We use Ascend 910B to test whether \sys{} can obtain useful optimization knowledge from the same five expert kernels on a non-NVIDIA platform.
To this end, we induce the library from expert implementations in \texttt{torch-npu}, \texttt{cann-ops} and \texttt{vllm-ascend}, using the same five operator categories and workload shapes as on NVIDIA. Of 109 candidate skills, 67 pass roundtrip admission, yielding a rate of $61.5\%$, compared with $72.2\%$ on H100 and $80.0\%$ on RTX PRO 6000 (\autoref{tab:roundtrip}). Reaching a correct naive kernel on Ascend also requires more deoptimization steps than on NVIDIA. These difficulties are consistent with weaker model familiarity with Ascend programming, reflecting more limited coverage in pretraining data. The one-time offline induction cost is \$197.

We then generate AscendC kernels for the five workloads from naive implementations using the induced skills. The generated kernels outperform the expert baselines on all five workloads, with speedups of $1.048\times$--$1.496\times$ (\autoref{tab:ascend-compare}).

\begin{table}[h]
  \centering
  \small
  \setlength{\tabcolsep}{4.5pt}
  \begin{tabular}{@{}lccccc@{}}
    \toprule
    \textbf{Time ($\mu$s)} & \textbf{GEMM} & \textbf{Conv} & \textbf{FMHA} & \textbf{GDN} & \textbf{Top-K} \\
    \midrule
    \sys{}     & 424.2 & 17.74 & 6518.3 & 3393.4 & 15.48 \\
    Expert Baseline  & 444.6 & 26.54 & 7531.0 & 4764.4 & 19.83 \\
    \midrule
    Speedup    & $1.048\times$ & $1.496\times$ & $1.155\times$ & $1.404\times$ & $1.281\times$ \\
    \bottomrule
  \end{tabular}
  \caption{Generated AscendC-kernel speedup over the platform-specific vendor reference.}
  \label{tab:ascend-compare}
\end{table}

These results suggest that \sys{} can build on the small set of expert kernels available early in a new hardware platform's development. Engineers typically prioritize core operators such as GEMM, and guided deoptimization allows \sys{} to recover the hardware-specific knowledge in these implementations and reuse it to optimize other operators.

\subsection{RQ4: Transfer to New Operators and Benchmark Suites}
\label{sec:exp-transfer}

\paragraph{Operator transfer.}
On H100 and Ascend 910B, we evaluate transfer to five new operators: KDA, Sparse Attention, Top-P, Fused Add-RMSNorm, and GQA. None contributed to skill induction. We compare \sys{} with the Codex baseline, which receives no conditioned skills.

\begin{figure*}[h]
\centering
\includegraphics[width=\textwidth]{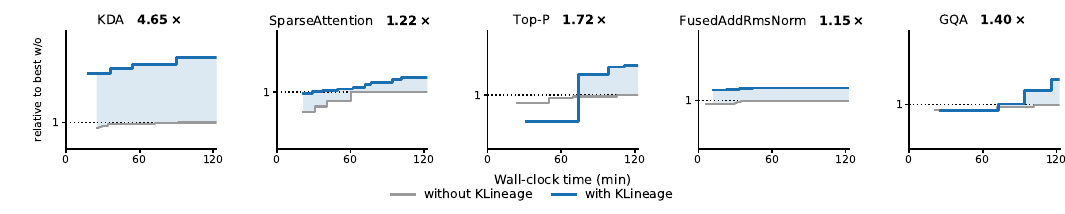}
\par\vspace{1pt}
{\scriptsize (a) H100 (SM90)\par}
\vspace{4pt}
\includegraphics[width=\textwidth]{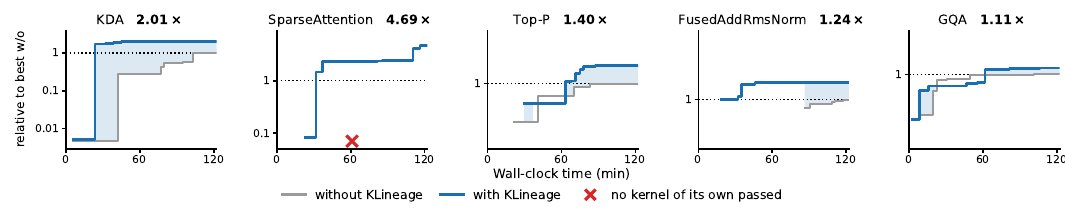}
\par\vspace{1pt}
{\scriptsize (b) Ascend 910B\par}
\caption{\sys improve generation across the five transfer targets on H100 and Ascend 910B within two hours.}
\label{fig:transfer-staircase}
\end{figure*}

At the end of the budget, \sys{} retains faster kernels on every target where both settings produce valid kernels. Relative to Codex baseline, the geometric-mean speedup is $1.73\times$ over all five H100 targets and $1.41\times$ over the four Ascend targets with valid results from both settings. On the remaining Ascend target, Sparse Attention, \sys{} produces a valid kernel while Codex baseline fails. \autoref{fig:transfer-staircase} shows the optimization trajectories, and Appendix~\ref{app:transfer-axes} reports the corresponding token and cost trajectories. These results demonstrate that the skill libraries induced by \sys{} transfer to kernel-generation tasks for operators not used during induction.

\paragraph{Optimizations recovered on Ascend.}
\autoref{tab:crossed} shows that \sys{} increases the total number of optimizations implemented across the five Ascend targets from 40 to 55, while preserving all optimizations found by the Codex baseline. Blue checkmarks indicate optimizations found only with \sys{}.

\begin{table*}[h]
\centering
\small
\setlength{\tabcolsep}{3.4pt}
\newcommand{\yes}{$\checkmark$}
\newcommand{\no}{\textcolor{black!35}{$\times$}}
\newcommand{\gain}{\textcolor{blue!70!black}{$\boldsymbol{\checkmark}$}}
\begin{tabular}{@{}l cc cc cc cc cc@{}}
\toprule
 & \multicolumn{2}{c}{\textbf{KDA}} & \multicolumn{2}{c}{\textbf{Top-P}}
 & \multicolumn{2}{c}{\textbf{GQA}} & \multicolumn{2}{c}{\textbf{Sparse Attn.}}
 & \multicolumn{2}{c}{\textbf{Add-RMSNorm}} \\
\cmidrule(lr){2-3}\cmidrule(lr){4-5}\cmidrule(lr){6-7}\cmidrule(lr){8-9}\cmidrule(lr){10-11}
\textbf{Optimization} & w/o & w/ & w/o & w/ & w/o & w/ & w/o & w/ & w/o & w/ \\
\midrule
Cube matmul for attention products   & \no & \no  & \no & \no   & \yes & \yes & \no & \gain & \no & \no   \\
Cube products split from Vector softmax & \no & \no & \no & \no & \yes & \yes & \no & \gain & \no & \no   \\
Work-aware launch partition          & \no & \no  & \no & \no   & \no & \gain & \no & \no   & \no & \gain \\
\addlinespace[1pt]
Two-slot / ping-pong staging         & \yes & \yes & \yes & \yes & \yes & \yes & \no & \gain & \yes & \yes \\
Prefetch ahead of compute            & \no & \gain & \no & \no   & \no & \no   & \no & \gain & \yes & \yes \\
Pack or transpose for downstream access & \no & \gain & \no & \no & \yes & \yes & \no & \no  & \no & \no   \\
Event-ordered overlap across streams  & \no & \no  & \no & \no   & \no & \gain & \no & \gain & \no & \no   \\
Workspace reuse across phases        & \no & \no  & \no & \no   & \no & \gain & \no & \gain & \no & \no   \\
L2 bypass for streaming writes       & \no & \no  & \no & \no   & \no & \no   & \no & \no   & \no & \gain \\
\addlinespace[1pt]
Row-max subtraction before exp       & \no & \no  & \no & \no   & \no & \gain & \yes & \yes & \no & \no   \\
GatherMask compaction of threshold band & \no & \no & \no & \gain & \no & \no  & \no & \no   & \no & \no   \\
\midrule
\multicolumn{11}{@{}l}{\textit{Nine further optimizations appear in both settings; the full audit is in \autoref{app:technique-matrix}.}} \\
\midrule
\textbf{Total (of 20)} & 7 & \textbf{9} & 8 & \textbf{9} & 10 & \textbf{14} & 7 & \textbf{13} & 8 & \textbf{10} \\
\bottomrule
\end{tabular}
\caption{\sys{} retains all 40 optimizations found without skills and adds 15 across the five Ascend transfer targets.}
\label{tab:crossed}
\end{table*}

\subsubsection{Generality to Broader Operators}
\label{sec:exp-kernelbench}

We test generality beyond the expert workloads on the $100$ Level~2 problems of KernelBench~\citep{kernelbench}, run on NVIDIA in CUDA and, as the Fusion category of MultiKernelBench~\citep{multikernelbench}, ported to AscendC on Ascend 910B. The two platforms therefore face the same $100$ problems rather than two different suites. Each platform's skill library remains unchanged, and none of these problems contributed to induction. Speedup is measured against \texttt{torch.compile} on NVIDIA and eager PyTorch on Ascend, then averaged over all valid solutions in each setting.

\autoref{tab:broader-operators} reports both evaluations. On NVIDIA, both settings produce correct kernels for all 100 problems; \sys{} raises mean speedup from $1.05\times$ to $1.32\times$ and the number of problems faster than \texttt{torch.compile} from 26 to 42. On Ascend, correctness rises from 68 to 100 problems, mean speedup from $0.63\times$ to $2.09\times$, and the number faster than eager from 22 to 93.

NVIDIA runs incur higher API costs because faster kernel execution and profiling allow more agent turns within the same wall-clock budget.

\begin{table}[ht]
  \centering
  \small
  \setlength{\tabcolsep}{6pt}
  \begin{tabular}{@{}lcccc@{}}
    \toprule
    \textbf{Method} & \textbf{Mean} & \textbf{Correct} & \textbf{Faster than} & \textbf{Cost} \\
                    & \textbf{speedup} & \textbf{(\%)}   & \textbf{reference (\%)}  & \textbf{(USD)} \\
    \midrule
    \multicolumn{5}{@{}l}{\textit{KernelBench Level 2 --- NVIDIA, CUDA}} \\
    without \sys{} & $1.05\times$ & 100 & 26 & 28.77 \\
    with \sys{} & $\mathbf{1.32\times}$ & 100 & \textbf{42} & 31.19 \\
    \midrule
    \multicolumn{5}{@{}l}{\textit{MultiKernelBench Fusion --- Ascend 910B, AscendC}} \\
    without \sys{} & $0.63\times$ & 68 & 22 & 21.92 \\
    with \sys{} & $\mathbf{2.09\times}$ & \textbf{100} & \textbf{93} & 24.05 \\
    \bottomrule
  \end{tabular}
  \caption{\sys{} improves performance on both broader operator benchmarks while producing correct kernels for all evaluated problems.}
  \label{tab:broader-operators}
\end{table}

\section{Conclusion}
\sys recovers the missing \emph{when} of GPU-kernel optimization from expert artifacts by walking them backward under validation-gated deoptimization and lifting accepted forward transitions into code-anchored, scope-tagged skills. Expert kernels become not only outputs to imitate but evidence from which the conditional structure of optimization can be externalized and reused.

\section*{AI use statement}
We used AI to construct conditioned skill memory for each of the three hardware platforms and to generate the final kernel code. We also used AI to polish the paper's writing. We reviewed the LLM-generated code and take responsibility for the final content of this paper.

\bibliography{custom}
\bibliographystyle{preprint}

\appendix

\section{Workload Details}
\label{app:workload-detail}

\autoref{tab:workload-detail} lists the shape, dtype and reference for the five expert kernels
and five transfer targets on each platform. Expert kernels are the sources from which
\sys{} induces skills; transfer targets contribute no lineage.

\begin{table*}[h]
  \centering
  \small
  \setlength{\tabcolsep}{5pt}
  \begin{tabular}{@{}llp{0.40\textwidth}l@{}}
    \toprule
    \textbf{Role} & \textbf{Kernel} & \textbf{Shape (dtype)} & \textbf{Reference} \\
    \midrule
    \multicolumn{4}{@{}l}{\textit{NVIDIA --- expert kernels}} \\
    source & GEMM   & $M{=}N{=}K{=}4096$ (BF16) & cuBLAS \\
    source & Conv2d & $N{=}8$, $C{=}64$, $H{=}W{=}56$, $F{=}128$, $K{=}3$, s${=}1$, p${=}1$ (FP16) & cuDNN \\
    source & FMHA   & batch $8$, heads $64$, seq $2048$, dim $128$ (FP16) & FlashInfer \\
    source & GDN    & $h_{qk}{=}16$, $h_v{=}48$, dim $128$, chunk $64$, seq $4096$ (BF16) & FlashInfer / FLA \\
    source & Top-K  & batch $64$, seq $4096$, top-$k{=}512$ (FP32) & FlashInfer \\
    \multicolumn{4}{@{}l}{\textit{NVIDIA --- transfer targets}} \\
    target & KDA               & batch $1$, tokens $4096$, heads $96$, dim $128$ (BF16) & FLA \\
    target & Sparse Attention  & queries $8192$, heads $128$, $2048$ sparse indices & FlashInfer \\
    target & Top-P             & batch size $32$, vocabulary $128256$ & FlashInfer \\
    target & Fused Add-RMSNorm & batch size $8192$, hidden $7168$ & Torch \\
    target & GQA               & tokens $16384$, $32$ query / $8$ KV heads & FlashInfer \\
    \midrule
    \multicolumn{4}{@{}l}{\textit{Ascend --- expert kernels}} \\
    source & GEMM   & $M{=}N{=}K{=}4096$ (BF16) & Torch-NPU  \\
    source & Conv2d & $N{=}8$, $C{=}64$, $H{=}W{=}56$, $F{=}128$, $K{=}3$, s${=}1$, p${=}1$ (FP16) & AscendC custom op \\
    source & FMHA   & batch $8$, heads $64$, seq $2048$, dim $128$ (FP16) & Torch-NPU  \\
    source & GDN    & $h_{qk}{=}16$, $h_v{=}48$, dim $128$, chunk $64$, seq $4096$ (BF16) & vLLM-Ascend \\
    source & Top-K  & batch $64$, seq $4096$, top-$k{=}512$ (FP32) & Torch-NPU  \\
    \multicolumn{4}{@{}l}{\textit{Ascend --- transfer targets}} \\
    target & KDA               & batch $1$, tokens $4096$, heads $96$, dim $128$ (BF16) & Torch-NPU \\
    target & Sparse Attention  & queries $8192$, heads $128$, $2048$ sparse indices & AscendC custom op \\
    target & Top-P             & batch size $32$, vocabulary $128256$ & Torch-NPU \\
    target & Fused Add-RMSNorm & batch size $8192$, hidden $7168$ & Torch-NPU \\
    target & GQA               & tokens $16384$, $32$ query / $8$ KV heads & AscendC custom op \\
    \bottomrule
  \end{tabular}
  \caption{The workload suite separates five expert kernels used for induction from five targets used for transfer on each platform.}
  \label{tab:workload-detail}
\end{table*}

\section{Source-Tier Reconstruction and Memory-Based Kernel Agents}
\label{sec:exp-cuda}

This appendix backs \autoref{fig:cuda-bars} with absolute latencies, cost trajectories and the
per-step record of one baseline run. AdaExplore~\citep{adaexplore} builds failure-rule memory with
MCTS-style exploration; AccelOpt~\citep{accelopt} maintains an LLM-summarized memory over generated
kernels. Both are re-run in our harness with their search policy and memory schema preserved
(\autoref{sec:appendix-baselines}). These numbers are a reconstruction check, not a transfer claim:
the operators here are the ones whose lineages the library was built from.

\section{Main-Tier Per-Instance Latencies}
\label{app:tables}

\autoref{tab:appendix-cuda-latencies} gives the absolute per-kernel latencies behind
\autoref{fig:cuda-bars}. ``FAIL'' indicates that the method did not produce a correct
kernel within the budget; $\dagger$ marks a post-hoc adapter audit rather than an
as-submitted valid kernel.

\begin{table*}[h]
  \centering
  \small
  \setlength{\tabcolsep}{6pt}
  \begin{tabular}{@{}llrrrr@{}}
    \toprule
    \textbf{Platform} & \textbf{Kernel} & \textbf{\sys{} (ms)} & \textbf{AdaExplore (ms)} & \textbf{AccelOpt (ms)} & \textbf{Reference (ms)} \\
    \midrule
    SM90  & GEMM   & 0.3130  & 0.3198   & 0.3190  & 0.3086 \\
    SM90  & Conv2d & 0.0157  & 0.0684   & 0.0658  & 0.0272 \\
    SM90  & FMHA   & 3.5183  & FAIL     & 5.1499  & 3.3121 \\
    SM90  & GDN    & 0.2439  & 17.7673  & FAIL    & 0.2816 \\
    SM90  & Top-K  & 0.0143  & 0.1338   & 0.1276  & 0.0160 \\
    \midrule
    SM120 & GEMM   & 0.3681  & 0.3896   & 0.3933  & 0.3688 \\
    SM120 & Conv2d & 0.0235  & 0.0760$^\dagger$ & 0.0822  & 0.0253 \\
    SM120 & FMHA   & 3.2151  & FAIL     & 3.6366  & 3.1932 \\
    SM120 & GDN    & 0.6348  & FAIL     & FAIL    & 0.6867 \\
    SM120 & Top-K  & 0.0131  & FAIL     & 0.0288  & 0.0169 \\
    \bottomrule
  \end{tabular}
  \caption{\sys{} produces correct kernels on all ten CUDA workloads, while AdaExplore and AccelOpt fail on four and two, respectively.}
  \label{tab:appendix-cuda-latencies}
\end{table*}

\paragraph{Expert-source comparison.}
\autoref{tab:expert-source-ablation} gives the per-workload reconstruction latencies for the comparison in \autoref{sec:exp-source-ablation}. The Gen. column denotes expert-source prompting, and slowdown is its latency divided by the corresponding \sys{} latency.

\begin{table}[t]
  \centering
  \small
  \setlength{\tabcolsep}{6pt}
  \begin{tabular}{@{}llrrr@{}}
    \toprule
    \textbf{Platform} & \textbf{Kernel} & \textbf{Gen. (ms)} & \textbf{\sys{} (ms)} & \textbf{Slowdown} \\
    \midrule
    SM90  & GEMM   & 0.4070  & 0.3130 & $1.30\times$ \\
    SM90  & Conv2d & 0.2273  & 0.0157 & $14.48\times$ \\
    SM90  & FMHA   & 27.7556 & 3.5183 & $7.89\times$ \\
    SM90  & GDN    & 25.6522 & 0.2439 & $105.18\times$ \\
    SM90  & Top-K  & 0.0211  & 0.0143 & $1.48\times$ \\
    \midrule
    SM120 & GEMM   & 0.4781  & 0.3681 & $1.30\times$ \\
    SM120 & Conv2d & 0.1369  & 0.0235 & $5.83\times$ \\
    SM120 & FMHA   & 11.8063 & 3.2151 & $3.67\times$ \\
    SM120 & GDN    & 23.7612 & 0.6348 & $37.43\times$ \\
    SM120 & Top-K  & 0.0147  & 0.0131 & $1.12\times$ \\
    \bottomrule
  \end{tabular}
  \caption{Conditioned skills outperform direct expert-source prompting on all ten CUDA workloads under the same \$10 budget.}
  \label{tab:expert-source-ablation}
\end{table}

\begin{figure*}[t]
\centering
\includegraphics[width=\textwidth]{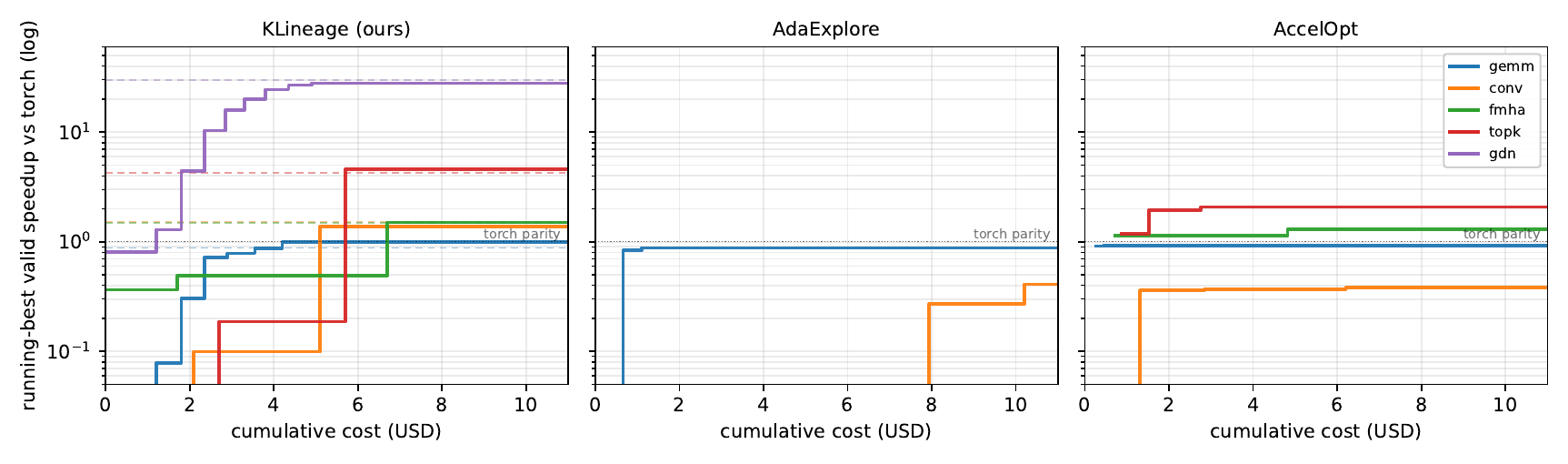}
\caption{\sys{} reaches its best SM120 kernels within \$4--\$7 of online optimization cost.}
\label{fig:budget-curves}
\end{figure*}

\paragraph{Cost over the budget.}
\autoref{fig:budget-curves} reads the same comparison as a function of cumulative LLM-API cost on
SM120. Loading the induced library costs around $\$0.7$ in retrieval and load --- the flat plateau
at the start of each \sys{} curve --- and is paid once per workload, since lineages and SkillCards
stay in a single prefix-cached chat session. After the load, \sys{} reaches its eventual best
between $\$4$ and $\$7$, well under the cap, and each step lines up with one named, verified skill
rather than a black-box submission.

\section{Baseline Execution and Harness Notes}
\label{sec:appendix-baselines}

We evaluate AdaExplore~\citep{adaexplore} and AccelOpt~\citep{accelopt} in the same external harness used for \sys. This harness standardizes the parts that determine the reported score---input shapes, correctness tests, profiling, vendor references, and cost accounting---while preserving each baseline's own search policy and memory representation. In other words, we do not replace either baseline with a \sys-style memory or planner; each method still decides what to submit using its original in-context state.

All three methods use Claude Opus 4.6 as the online materialization backbone, the per-workload \$$10$ optimization-cost budget, the same compile/correctness/profile gate, and the same vendor reference per platform (\autoref{sec:exp-setup}, \autoref{tab:workload-detail}). We use a dollar budget rather than a fixed submission count because the methods differ in session structure and cache reuse: \sys keeps one prefix-cached chat session per workload, whereas the baselines either restart sessions per submission or use looser caching. The goal is therefore to compare optimization usefulness under the same online cost, not to measure each method's asymptotic ceiling under an unconstrained search horizon.

\paragraph{AdaExplore.}
For AdaExplore, we use the released failure-rule memory, failure-tag extraction, MCTS-style exploration, action vocabulary, and per-step prompt templates. The failure-rule memory is reset at the start of each workload, matching the published per-task setting and preventing rules from one workload from leaking into another. The only harness-level changes are that submissions are routed through our evaluator and budget meter, so that compile status, correctness, latency, and LLM cost are measured consistently with \sys and AccelOpt. We do not modify AdaExplore's search policy, rule-extraction prompt, or memory schema.

\paragraph{Non-kernel wrapper outputs.}
A practical issue arises in some AdaExplore runs: several submissions reach the evaluator as wrapper programs rather than executable generated kernels. The agent emits a Triton kernel definition, but the surrounding \texttt{ModelNew} catches the kernel-call failure and returns the PyTorch reference. Such submissions can pass correctness, and may even appear fast, but the measured program is then the fallback reference path rather than the generated kernel. We verified this by replacing the fallback return path with a \texttt{raise}; under the same workload interface, these submissions no longer pass.

We therefore exclude fallback-wrapper scores from the generated-kernel comparison. To avoid under-crediting AdaExplore, we additionally perform a post-hoc audit for SM120 Conv2d by wiring a thin adapter that directly calls the underlying generated Conv2d kernels with the case-side $(x,w)$ tensors. This audit recovers two real kernels, but they are much slower than cuDNN (best about $0.41\times$ vs.\ torch and $0.33\times$ vs.\ cuDNN near the end of the run), and the adapter itself is not produced by AdaExplore within its workflow. We mark this number with $\dagger$ in \autoref{tab:appendix-cuda-latencies}: it credits the real audited kernel rather than the wrapper's fallback score. We do not otherwise repair AdaExplore submissions or introduce task-specific adapters, since such harness-level repair is outside the published AdaExplore loop and the shared budget.

\paragraph{AccelOpt.}
For AccelOpt, we use the released iterative refine-with-summary loop and its per-step memory-summary prompt. The memory state is initialized empty for each workload, matching the published per-task setting. As with AdaExplore, submissions are routed through our evaluator and budget meter, but the baseline's memory remains an LLM-summarized rolling state over its own forward submissions. We do not introduce SkillCards, lineage retrieval, expert-derived transitions, or any \sys-specific memory schema into AccelOpt.

\paragraph{Shared scoring protocol.}
All three methods share the online backbone, the per-workload cost budget, the evaluator (\texttt{triton.testing.do\_bench} with \texttt{warmup=25} and \texttt{rep=100}, both in milliseconds), the correctness check (three random seeds), the speedup definition ($T_{\mathrm{ref}}/T_{\mathrm{gen}}$), and the run count ($5$ runs per workload, reported as success-only geometric mean). The vendor reference is fixed per platform and workload: cuBLAS for GEMM, cuDNN for Conv2d, FlashInfer for FMHA and Top-K, and FlashInfer or FLA for GDN (\autoref{tab:workload-detail}). Thus, the comparison keeps the model, evaluator, budget, and references fixed; what varies is the optimization memory each method carries between submissions.

\paragraph{Scope of the comparison.}
This comparison evaluates fixed-cost online optimization, not unlimited-horizon search. Search-heavy methods may improve with additional budget, and \sys could also use additional budget for repair and local exploration. Our goal in the main comparison is to ask how useful each memory representation is under the same online materialization cost and validation gate. Per-workload absolute latencies for both baselines are listed alongside \sys in \autoref{tab:appendix-cuda-latencies}.

\begin{figure*}[h]
\centering
\includegraphics[width=0.78\textwidth]{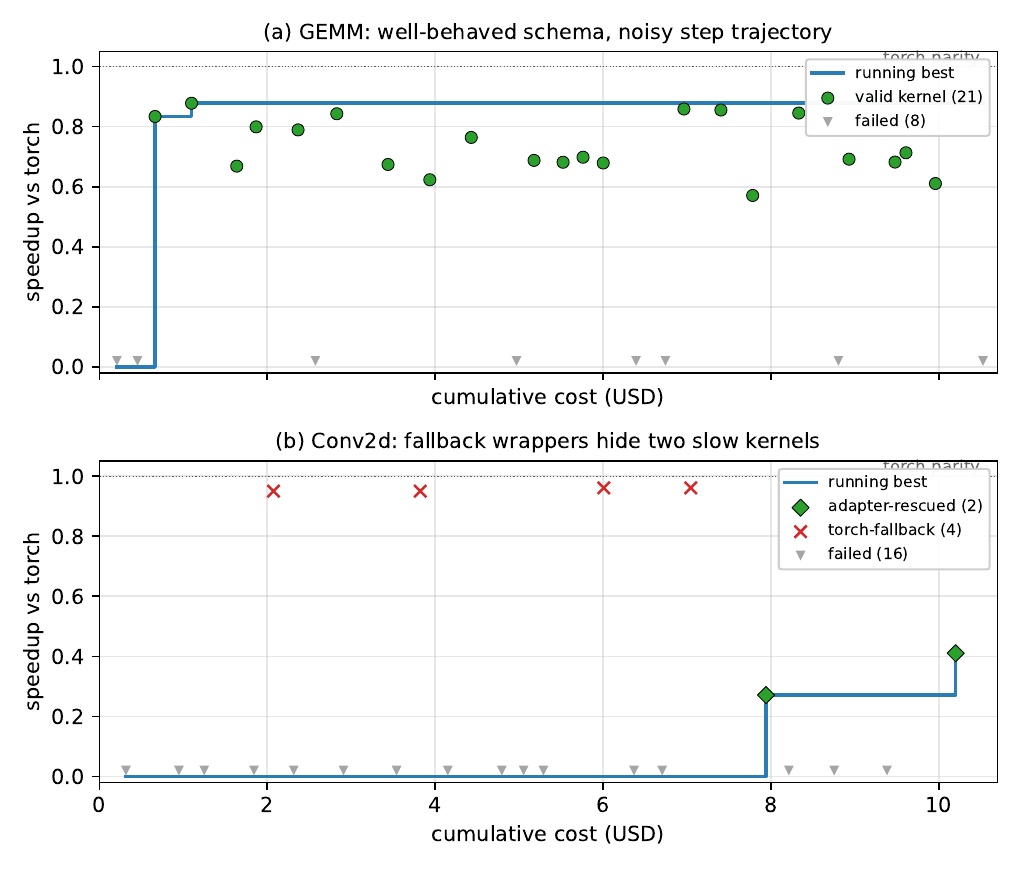}
\caption{AdaExplore's SM120 trajectories show repeated regressions on GEMM and frequent failures or fallback wrappers on Conv2d.}
\label{fig:ada-detail}
\end{figure*}

\paragraph{What an AdaExplore trajectory looks like at the step level.}
\autoref{fig:ada-detail} shows the per-step trajectory for two of the AdaExplore SM120 cases. Panel (a) is GEMM, where $21$ valid kernels are produced over the $\$10$ budget at speedups ranging from $0.57\times$ to $0.88\times$, plus $8$ failed steps; the running-best reaches $0.88\times$ at $\$1.1$ and stays flat afterwards, and most subsequent valid kernels regress below the running-best rather than improve it. Panel (b) is Conv2d, where most steps fail outright ($16$ failed) or emit non-kernel wrappers that compute the right answer through PyTorch ($6$ flagged in the raw trajectory). A thin post-hoc adapter can benchmark two underlying kernels as real Conv2d kernels, reaching about $0.41\times$ near the end of the run, while the remaining wrappers are excluded. The two panels jointly motivate why we report running-best over a wrapper-filtered step set rather than picking the wrapper's reported speedup at face value: the latter would have credited the Conv2d wrappers with about $0.96\times$ on this case.

Green circles denote valid submissions, green diamonds denote adapter-audited kernels, red crosses denote excluded fallback wrappers, and grey triangles denote failed steps. The blue line tracks the best wrapper-filtered speedup, including the audited Conv2d kernels.

\section{CUDA and Ascend Trajectories against Tokens and Cost}
\label{app:transfer-axes}

\begin{figure*}[h]
\centering

\includegraphics[width=\textwidth]{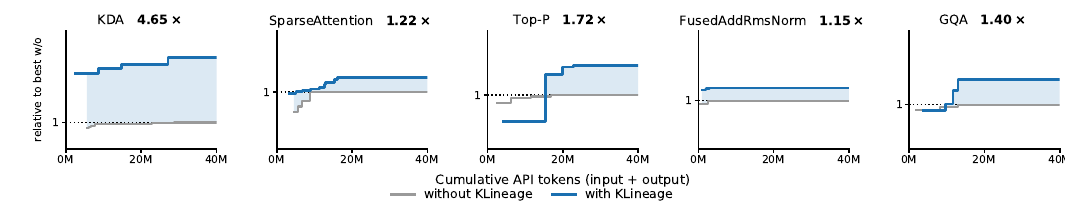}

\vspace{4pt}

\includegraphics[width=\textwidth]{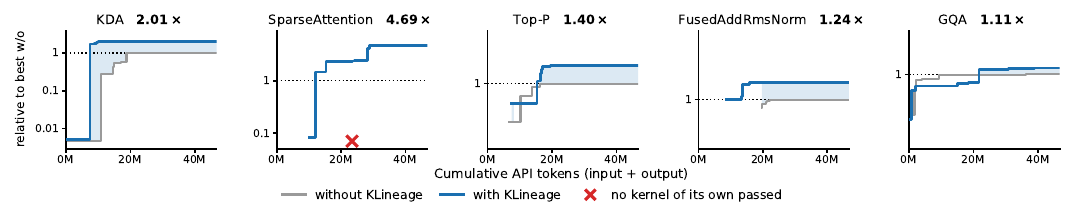}

\caption{Token-based trajectories show the transfer gains from conditioned skills on both CUDA and Ascend.}
\label{fig:transfer-tokens}
\end{figure*}

\begin{figure*}[h]
\centering

\includegraphics[width=\textwidth]{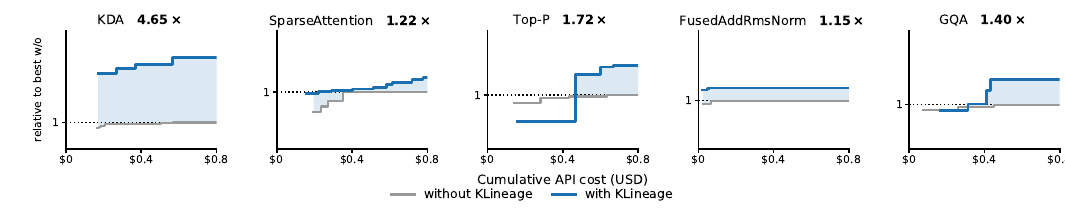}

\vspace{4pt}

\includegraphics[width=\textwidth]{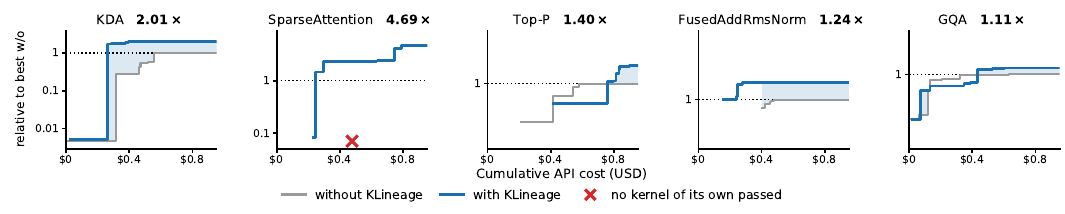}

\caption{Transfer trajectories against API cost show faster final kernels with conditioned skills on both platforms.}
\label{fig:transfer-cost}
\end{figure*}

\autoref{fig:transfer-staircase} shows the CUDA and Ascend transfer trajectories against wall-clock time in the top and bottom panels, respectively. Each operator is normalized by the best kernel reached without \sys{}, except Ascend Sparse Attention, which uses the eager reference because the run without \sys{} produced no valid kernel.
To complement the time-based view, \autoref{fig:transfer-tokens} and \autoref{fig:transfer-cost} plot the same runs against cumulative API tokens and monetary cost, respectively, with CUDA above Ascend in each figure. Token counts include input and output tokens, including cache hits; normalization and shading follow the corresponding wall-clock plots.
These axes better capture agent-side optimization effort when a session is idle, interrupted, or resumed, where wall-clock time may not faithfully reflect the amount of model interaction.

For the cost analysis, we compute API cost using the DeepSeek V4.1-Flash peak rates: \$$0.30$ per million cache-miss input tokens, \$$0.006$ per million cache-hit input tokens, and \$$1.20$ per million output tokens, charged per API call.
This provides an upper bound on the actual cost of these runs, since only the Sparse Attention pair was executed during the peak-pricing window.

\section{AscendC Technique Matrix}
\label{app:technique-matrix}

\autoref{tab:technique-matrix} audits the final submission of each setting on the five Ascend transfer
targets for the optimization mechanisms visible in its source. A mechanism is counted when its
implementation is present in the selected bundle; the audit describes code, not an isolated
performance effect, and it does not establish that any one mechanism caused a speedup. The setting with
\sys{} contains everything the setting without it found on every target: no mechanism is lost, and fifteen
are added. Blue checkmarks identify optimizations present only in the \sys{} setting.

\begin{table*}[h]
\centering
\small
\setlength{\tabcolsep}{2.8pt}
\newcommand{\yes}{$\checkmark$}
\newcommand{\no}{\textcolor{black!35}{$\times$}}
\newcommand{\gain}{\textcolor{blue!70!black}{$\boldsymbol{\checkmark}$}}
\begin{tabular}{@{}l cc cc cc cc cc@{}}
\toprule
 & \multicolumn{2}{c}{\textbf{KDA}} & \multicolumn{2}{c}{\textbf{Top-P}}
 & \multicolumn{2}{c}{\textbf{GQA}} & \multicolumn{2}{c}{\textbf{Sparse Attn.}}
 & \multicolumn{2}{c}{\textbf{Add-RMSNorm}} \\
\cmidrule(lr){2-3}\cmidrule(lr){4-5}\cmidrule(lr){6-7}\cmidrule(lr){8-9}\cmidrule(lr){10-11}
\textbf{Optimization} & w/o & w/ & w/o & w/ & w/o & w/ & w/o & w/ & w/o & w/ \\
\midrule
\addlinespace[1pt]
\multicolumn{11}{@{}l}{\textit{Compute mapping}} \\
AIV vector arithmetic and reductions & \yes & \yes & \yes & \yes & \yes & \yes & \yes & \yes & \yes & \yes \\
Cube matmul for attention products & \no & \no & \no & \no & \yes & \yes & \no & \gain & \no & \no \\
Cube products split from Vector softmax & \no & \no & \no & \no & \yes & \yes & \no & \gain & \no & \no \\
Core-parallel row / token / head-slice split & \yes & \yes & \yes & \yes & \yes & \yes & \yes & \yes & \yes & \yes \\
Work-aware launch partition & \no & \no & \no & \no & \no & \gain & \no & \no & \no & \gain \\
\addlinespace[1pt]
\multicolumn{11}{@{}l}{\textit{Memory and pipeline}} \\
GM-to-UB staging (DataCopy, LocalTensor) & \yes & \yes & \yes & \yes & \yes & \yes & \yes & \yes & \yes & \yes \\
Two-slot / ping-pong staging & \yes & \yes & \yes & \yes & \yes & \yes & \no & \gain & \yes & \yes \\
Prefetch ahead of compute & \no & \gain & \no & \no & \no & \no & \no & \gain & \yes & \yes \\
Pack or transpose for downstream access & \no & \gain & \no & \no & \yes & \yes & \no & \no & \no & \no \\
Event-ordered overlap across streams & \no & \no & \no & \no & \no & \gain & \no & \gain & \no & \no \\
Workspace reuse across phases & \no & \no & \no & \no & \no & \gain & \no & \gain & \no & \no \\
L2 bypass for streaming writes & \no & \no & \no & \no & \no & \no & \no & \no & \no & \gain \\
Explicit DMA / Vector / Cube synchronization & \yes & \yes & \yes & \yes & \yes & \yes & \yes & \yes & \yes & \yes \\
\addlinespace[1pt]
\multicolumn{11}{@{}l}{\textit{Numerics and data semantics}} \\
FP32 vector intermediates & \yes & \yes & \yes & \yes & \yes & \yes & \yes & \yes & \yes & \yes \\
Row-max subtraction before exp & \no & \no & \no & \no & \no & \gain & \yes & \yes & \no & \no \\
Compensated FP32 sum in cutoff search & \no & \no & \yes & \yes & \no & \no & \no & \no & \no & \no \\
Invalid-index / causal / tail masking & \no & \no & \yes & \yes & \yes & \yes & \yes & \yes & \no & \no \\
GatherMask compaction of threshold band & \no & \no & \no & \gain & \no & \no & \no & \no & \no & \no \\
\addlinespace[1pt]
\multicolumn{11}{@{}l}{\textit{Operator-specific reuse}} \\
Recurrent state kept in local memory & \yes & \yes & \no & \no & \no & \no & \no & \no & \no & \no \\
Fused row sum reused for both outputs & \no & \no & \no & \no & \no & \no & \no & \no & \yes & \yes \\
\midrule
\textbf{Total} & 7 & \textbf{9} & 8 & \textbf{9} & 10 & \textbf{14} & 7 & \textbf{13} & 8 & \textbf{10} \\
\bottomrule
\end{tabular}
\caption{\sys{} preserves every optimization found by expert-source prompting and adds further optimizations on all five Ascend targets.}
\label{tab:technique-matrix}
\end{table*}

\section{Case Study: Gated Delta Net}
\label{sec:case-gdn}

We use GDN to inspect what a lineage records beyond a final latency number. The operator decomposes into three FlashQLA sub-kernels (\texttt{cumsum}, \texttt{fused\_fwd}, \texttt{kkt\_solve}) with different optimization profiles. \autoref{fig:gdn-trace} shows the deoptimization chains; its backward edges remove optimizations, and \sys{} stores the inverse directions as forward skill candidates. The admitted SkillCards are listed with their verification evidence in \autoref{tab:gdn-skills} (App.~\ref{app:gdn-skills}).

The point of the trace is not only that these optimizations exist, but that the intermediate code states make them learnable. A memory item that is too concrete copies one expert surface and is hard to reuse; one that is too abstract names an optimization but gives the agent no actionable edit. \sys{} first validates each concrete before/after transition by roundtrip execution, then aggregates validated transitions into intents with anchors, carriers and preconditions. This is why the resulting skill can remain actionable on code while still transferring across cases and languages.

\begin{figure}[!htbp]
\centering
\scriptsize
\resizebox{\columnwidth}{!}{%
\begin{tikzpicture}[
  every node/.style={font=\scriptsize},
  state/.style={draw,rounded corners,align=center,minimum width=1.30cm,minimum height=0.70cm,inner sep=1.5pt},
  action/.style={->,>=Latex,thick}
]
\node[anchor=west,font=\scriptsize\itshape] at (0.0,0.65) {\texttt{cumsum}};
\node[state] (c0) at (0.0,0)  {expert\\$0.011$\,ms};
\node[state] (c1) at (1.55,0) {$-$vec.\,load\\$0.012$\,ms};
\node[state] (c2) at (3.10,0) {$-$shuffle\\$0.012$\,ms};
\node[state] (c3) at (4.65,0) {$-$smem\\transpose\\$0.034$\,ms};
\node[state] (c4) at (6.20,0) {naive\\$0.035$\,ms};
\draw[action] (c0) -- (c1); \draw[action] (c1) -- (c2);
\draw[action] (c2) -- (c3); \draw[action] (c3) -- (c4);
\node[anchor=west,font=\scriptsize\itshape] at (0.0,-0.55) {\texttt{fused\_fwd}};
\node[state] (f0) at (0.0,-1.20)  {expert\\$0.549$\,ms};
\node[state] (f1) at (1.55,-1.20) {$-$L2\,raster\\$0.579$\,ms};
\node[state] (f2) at (3.10,-1.20) {$-$MMA\\$0.582$\,ms};
\node[state] (f3) at (4.65,-1.20) {$-$swizzle\\$19.65$\,ms};
\node[state] (f4) at (6.20,-1.20) {$-$phase\\decomp.\\$3017$\,ms};
\node[state] (f5) at (7.75,-1.20) {naive\\$6056$\,ms};
\draw[action] (f0) -- (f1); \draw[action] (f1) -- (f2);
\draw[action] (f2) -- (f3); \draw[action] (f3) -- (f4);
\draw[action] (f4) -- (f5);
\node[anchor=west,font=\scriptsize\itshape] at (0.0,-1.75) {\texttt{kkt\_solve}};
\node[state] (k0) at (0.0,-2.40)  {expert\\$0.122$\,ms};
\node[state] (k1) at (1.55,-2.40) {$-$warp\\spec.\\$0.122$\,ms};
\node[state] (k2) at (3.10,-2.40) {$-$MMA/TMA\\$1.163$\,ms};
\node[state] (k3) at (4.65,-2.40) {naive\\$18.5$\,ms};
\draw[action] (k0) -- (k1); \draw[action] (k1) -- (k2);
\draw[action] (k2) -- (k3);
\end{tikzpicture}
}
\caption{GDN deoptimization identifies shared-memory transpose, phase decomposition, and MMA/TMA layout as dominant optimizations in its three sub-kernels.}
\label{fig:gdn-trace}
\end{figure}

\paragraph{Three sub-kernels, three dominant skills.}
The largest drop in each trace isolates a different reusable intent. For \texttt{cumsum}, the dominant skill is a \emph{shared-memory transpose}: the scan axis is strided in input memory, and transposing each chunk into $[\text{head},\text{seq}]$ before scanning cuts $0.034{\to}0.012$\,ms, while vectorized access, warp-shuffle scan and intra-block parallelism contribute little at this shape. For \texttt{fused\_fwd}, it is \texttt{gemm\_phase\_decomposition}, which restructures per-token $O(C^3D)$ work into cooperative $O(C^2(D{+}VT))$ GEMM phases and accounts for the largest single improvement in the chain; the remaining MMA/swizzle skill matters only once the phase structure is in place. For \texttt{kkt\_solve}, cooperative threading pays before the MMA/TMA intrinsic skill becomes profitable.

The \emph{conditional dependency} is the point: on \texttt{cumsum} the layout transpose makes the on-chip scan worth doing at all; on \texttt{fused\_fwd} phase decomposition is a precondition for the tensor-core swizzle to matter; on \texttt{kkt\_solve} the cooperative-threading layout has to be in place before the MMA/TMA intrinsics fit. Each of these is a precondition a card records and a label does not.

\section{GDN SkillCards (Case Study)}
\label{app:gdn-skills}

\autoref{tab:gdn-skills} lists the admitted SkillCards from the GDN case study
(\autoref{sec:case-gdn}). Each row is one admitted skill; the verification-evidence column is the
speedup recorded in the held-out roundtrip trial that admitted the card.

\begin{table}[h]
  \centering
  \resizebox{\columnwidth}{!}{%
  \begin{tabular}{@{}lll@{}}
  \toprule
  \textbf{Skill} & \textbf{Source sub-kernel} & \textbf{Verification evidence} \\
  \midrule
  \texttt{vectorized\_global\_load\_store}     & \texttt{cumsum}     & $3.9\times$ over naive  \\
  \texttt{warp\_shuffle\_scan}                 & \texttt{cumsum}     & $2.7\times$ over naive  \\
  \texttt{smem\_transpose}                     & \texttt{cumsum}     & $2.9\times$ over naive  \\
  \texttt{multi\_threaded\_block\_parallelism} & \texttt{cumsum}     & $1.1\times$ over naive  \\
  \texttt{mma\_ldmatrix\_stmatrix\_swizzle}    & \texttt{fused\_fwd} & $34\times$ over scalar  \\
  \texttt{gemm\_phase\_decomposition}          & \texttt{fused\_fwd} & $194\times$ over naive  \\
  \texttt{multi\_threaded\_block\_parallelism} & \texttt{fused\_fwd} & $1.6\times$ over naive  \\
  \texttt{mma\_ldmatrix\_stmatrix\_swizzle\_tma} & \texttt{kkt\_solve} & $16\times$ over scalar  \\
  \texttt{multi\_threaded\_block\_parallelism} & \texttt{kkt\_solve} & $16\times$ over naive   \\
  \bottomrule
  \end{tabular}%
  }
  \caption{GDN's three sub-kernels yield verified skills for memory layout, parallelism, and matrix computation.}
  \label{tab:gdn-skills}
\end{table}

\end{document}